\documentclass[journal]{IEEEtran}   
\usepackage{algorithm}
\usepackage{algorithmic}

\usepackage[table,xcdraw]{xcolor}
\usepackage{nomencl}
\makenomenclature
\usepackage{amssymb}
\usepackage{arydshln}
\usepackage{indentfirst}
\usepackage{lipsum}
\usepackage{tabularx}
\usepackage{array}
\usepackage[noadjust]{cite}
\usepackage{epsfig}
\usepackage{physics}
\usepackage{hhline}
\usepackage{amssymb,amsmath,color,graphicx, amsbsy}
\usepackage{algorithm}
\usepackage{amsfonts}
\usepackage{epstopdf}
\usepackage{multirow}
\usepackage{verbatim}
\usepackage{mathtools}
\usepackage{nomencl}
\usepackage{graphicx}
\usepackage{xcolor}
 \usepackage{soul}
\usepackage{lipsum,multicol}
\usepackage{multicol}
\usepackage{dblfloatfix}
\usepackage{float}
\usepackage{cuted}
\graphicspath{ {./Images/}}
\usepackage{multirow}
\edef\oldtt{\ttdefault}
\usepackage[scaled]{beramono}
\usepackage[T1]{fontenc}
\renewcommand{\ttdefault}{\oldtt}

\usepackage{etoolbox}

\usepackage{multirow}
\usepackage[caption=false]{subfig}

\begin{document}


\title{\huge GraphPMU: Event Clustering via Graph Representation Learning Using Locationally-Scarce Distribution-Level Fundamental and Harmonic PMU Measurements}

\author{Armin Aligholian,~\IEEEmembership{Student Member, IEEE}, Hamed Mohsenian-Rad,~\IEEEmembership{Fellow, IEEE}
 \thanks{A. Aligholian and H. Mohsenian-Rad are with
the University of California, Riverside, CA, USA. The corresponding author is H. Mohsenian-Rad. 
}

\vspace{-0.8cm}}

\maketitle

\begin{abstract}
This paper is concerned with the complex task of identifying the \emph{type} and \emph{cause} of the events that are captured by distribution-level phasor measurement units (D-PMUs) in order to enhance situational awareness in power distribution systems. Our goal is to address two fundamental challenges in this field: a) \emph{scarcity in measurement locations} due to the high cost of purchasing, installing, and streaming data from D-PMUs; b) \emph{limited prior knowledge about the event signatures} due to the fact that the events are diverse, infrequent, and inherently unscheduled.
To tackle these challenges, we propose an \emph{unsupervised graph-representation learning} method, called GraphPMU, to significantly improve the performance in event clustering under \emph{locationally-scarce data availability} by proposing the following two new directions: 1) using the \emph{topological information} about the \emph{relative location} of the few available phasor measurement units on the graph of the power distribution network; 2) utilizing not only the commonly used \emph{fundamental} phasor measurements, bus also the less explored \emph{harmonic} phasor measurements in the process of analyzing the signatures of various events.
Through a detailed analysis of several case studies, we show that GraphPMU can highly outperform the prevalent methods in the literature. 

\vspace{0.25cm}	

\textit{\textbf{Keywords}:} D-PMU, H-PMU, locationally  scarce measurements, feeder topology, graph representation learning, 
event clustering. 
\end{abstract}

\vspace{-0.4cm}

\section{Introduction} \label{sec:Introduction} 
\subsection{Motivations and Challenges}
\label{subsec: motivation}

Modern power distribution systems consist of various elements
, including the utility equipment and devices, such as capacitor banks, 
transformer tap changers, and protection devices, as well as customer devices, such as different types of loads, plugged-in electric vehicles, renewable generation units, and other distributed energy resources. The typical and benign operation (i.e., switching on and switching off) as well as mis-operation of these elements can create different 
types of \emph{events} in the system. Capturing and understanding these events is crucial to achieve situational awareness about the operation of the power distribution systems and their components \cite{Hamed_book}.

Accordingly, a growing sub-area in the field of smart grids has emerged recently that focuses on the analysis of various events in power distribution systems. Great attention has been devoted to using data from distribution-level phasor measurement units (D-PMU), a.k.a micro-PMUs \cite{hamedreview}. D-PMUs provide synchronized three-phase measurements of the \emph{fundamental} phasors of voltage and current on power distribution systems. The very high reporting rate of D-PMUs, makes it possible to capture and analyze a wide range of informative events. 

Meanwhile, the field of smart grid sensors has continued to grow also in the area of measuring \emph{harmonic} synchrophasors, which can be obtained from another new class of smart grid sensors, namely harmonic phasor measurement units (H-PMUs) \cite{Three_piece, Fast, class, FFT}. In fact, it is envisioned that D-PMUs and H-PMUs my soon converge; such that D-PMUs can provide both fundamental and harmonic synchrophasors, as needed \cite{class, Hamed_book}. 

When we use data from D-PMUs to analyze events, there are at least two major challenges that we need to address: 
\begin{enumerate}
    \item \emph{Scarcity in Measurement Locations}: Due to the high cost of D-PMUs, including purchase, installation, and data streaming, it is typical that a power distribution feeder is equipped with only a small number of D-PMUs. Hence, although the available D-PMUs provide synchronized phasor measurements at high resolution, such high resolution data availability is \emph{limited} to only a few locations on the power distribution feeder.
    
    \vspace{0.05cm}
    
    \item \emph{Limited Prior Knowledge about Event Signatures}: Analysis of events in power distribution systems is difficult, because: 1) such events are diverse and many of them are infrequent; 2) most events are inherently unscheduled; therefore, we do not know when to expect them; 3) the cause of many events and hence the characteristics of their signatures are not known in advance \cite{arminjournal}.

\end{enumerate}

 \vspace{0.05cm}


The key to address the second issue is to conduct  \emph{unsupervised} learning to identify the \emph{type} and \emph{cause} of the events. This can be done through \emph{event clustering}. The goal in event clustering is to identify the type of the events with minimal prior knowledge, i.e., without prior labeling of the events. 

The key to address the first issue is to make use of any available \emph{contextual information}, to enhance our ability to do event clustering when we face locational scarcity in data availability. In this paper, the primary contextual information is the graph of the topology of the power distribution network.



\vspace{-0.1cm}

\subsection{Summary of Technical Contributions}

In this paper, we seek to significantly improve the performance in event clustering under locationally-scarce data availability by taking the following two
new directions:

     \vspace{0.05cm}

\begin{enumerate}
    \item Using the information about the \emph{relative location} of the phasor measurement units on the network topology, notwithstanding the fact that such locations are scarce.
    
     \vspace{0.05cm}
    
    \item Using not only the \emph{fundamental} phasor measurements, that are commonly used in the literature in this field, but also the \emph{harmonic} phasor measurements in the process of analyzing the signatures of various events.
\end{enumerate}

     \vspace{0.05cm}

Both of the above directions are ways of incorporating new contextual information to the task of event clustering. The first one takes into account the location of sensors. The second one takes into account the changes in the harmonics that are caused by each event, as seen at the \emph{same} existing sensor locations. 
    




\vspace{0.05cm}

In addition to taking the above new approaches, the methodologies that are developed in this paper to implement these new approaches carry new technical contributions, as listed below:


\begin{itemize}

\item A new unsupervised two-step \emph{spatio-temporal} feature learning method is developed based on Graph Neural Networks (GNN) and Auto Encoder Decoder (AED) to capture the locational and temporal information of the sensors on the distribution network. The time series of the measurements at each bus are transferred to a lower dimension latent space. 
Accordingly, a graph learning method is implemented to the obtained embedding vectors to extract the \emph{topology-related features} for event clustering. To the best of our knowledge, this is the first time that a physics-aware graph learning method is used for utilizing D-PMU (or H-PMU) data in event clustering.

\vspace{0.05cm}

\item A graph-level representation learning is developed which uses \emph{local-global mutual information maximization} to learn the structural connection of the event data with its node-level representation. To extract the shared information between graph-level and node-level embedding that is sensitive to the graph topology, a \emph{negative graph sampling} based on a random network tree structure is proposed. This makes the proposed GNN more aware of the system topology, by encoding aspects of the data that are shared across different local nodes, and proper adversarial sampling for mutual information estimation. 

\vspace{0.05cm}

\item 
Incorporating the 3rd and the 5th harmonic phasor measurements along with the fundamental phasor measurements into the above aforementioned design. This is done by training a separate AED module is trained for each individual harmonic order. Then, the new aggregated vectors are used as additional input to the proposed graph learning process in order to capture the underlying locational patterns for each event, by taking into account both fundamental and harmonic phasor measurements. 
\end{itemize}

\vspace{0.05cm}

We refer to the proposed new design as GraphPMU. The results from various case studies show that GraphPMU leads to significant improvements in the accuracy of event clustering, despite the locational scarcity in the available phasor measurements. 
Importantly, even if we do \emph{not} use the harmonic phasor measurements, the mere use of topological information in the proposed graph learning paradigm can highly improve the performance in event clustering, specifically for the case of small events. 
The case studies in this paper include comparison with the state of the art clustering methods as well as with a variety of the GraphPMU implementation options, with different structures to show the importance of system topology, local-global feature learning, and number of  sensors. 

\subsection{Literature Review}
The literature on data-driven event types analysis in distribution-level phasor measurements can be generally divided into two categories. First, there are studies that use supervised learning, i.e. event classification, such as those in \cite{yuan2021learning,dynamicmonitoring,situationalawr}. They require prior labeling of the events in the training data set, which may not be doable in practice. 

The second group are the studies that use unsupervised learning, i.e., they attempt to cluster the events by grouping their distinctive characteristics. 
In \cite{vsagcluster}, \emph{k}-means clustering and Ward's clustering are proposed to cluster voltage sag events. 
In \cite{faultcluster}, an unsupervised clustering method is proposed for some specific faults; such as single-line-to-ground versus line-to-line faults.
In \cite{arminjournal}, an unsupervised event detection and clustering method is proposed, which requires solving a mixed integer program. Importantly, the study in \cite{arminjournal} is limited to the analysis of phasor measurements from only one D-PMU. It is inherently unrelated to the idea of taking into account the topological information of D-PMUs in the clustering task. 

None of the unsupervised learning methods in \cite{arminjournal,vsagcluster,faultcluster} takes into account the information on the topology of the power distribution network. Furthermore, none of them takes into account the availability of harmonic phasor measurements in addition to the fundamental phasor measurements.


To consider the network topology in the process of event clustering, one plausible way is to do graph-based analysis. Graph theory and more generally graph-based analysis have been used in power systems, such as for event detection \cite{ma2019hierarchical},  event location identification \cite{pandey2020real}, data recovery and prediction \cite{james2019synchrophasor}, 
and to study power system dynamics \cite{zhao2022structure}.


Recent literature also includes the use of GNNs, to
address some prevalent power system issues, including the analysis of events. In \cite{yuan2022learning}, a \emph{supervised} GNN-based method is proposed for event classification in power transmission systems. No knowledge about the topology of the power transmission network is assumed to be available; therefore, full connectivity is assumed in the graph-level analysis. The events are labeled based on the data of voltage and frequency. In \cite{luo2021data}, the authors used Graph Convolutional Network (GCN) for short term voltage stability assessment.
Importantly, neither of the studies in \cite{luo2021data} or \cite{yuan2022learning} consider the issue of locational scarcity among the sensors within a known network topology. None of them also considers using harmonic phasor measurements. 

Finally, there is a rich literature on the analysis of power quality events using measurements related to harmonics. The focus is usually on the analysis of waveform measurements, such as in \cite{alam2020classification, wilson2020automated, ge2020deep}. For example, in \cite{ge2020deep}, the authors proposed an AED to extract the features for clustering the daily variations in steady-state voltage harmonics. Interestingly, while we \emph{do} use H-PMU measurements, our focus is \emph{not} on the typical analysis of steady-state harmonics. Instead, we use the harmonic phasor measurements in addition to the fundamental phasor measurements to better capture the \emph{distinctive transient signatures} in various events in power distribution systems under locationally-scarce phasor measurements. Furthermore, the prior studies in this field, including those in \cite{alam2020classification, wilson2020automated, ge2020deep}, do \emph{not} consider using the information about the network topology.

\section{Topology-Based Representation Learning} \label{sec:topology}



\begin{figure}[t]
\begin{center}
\includegraphics[trim={1cm 0.5cm 0.5cm 0.7cm},clip,scale=0.58]{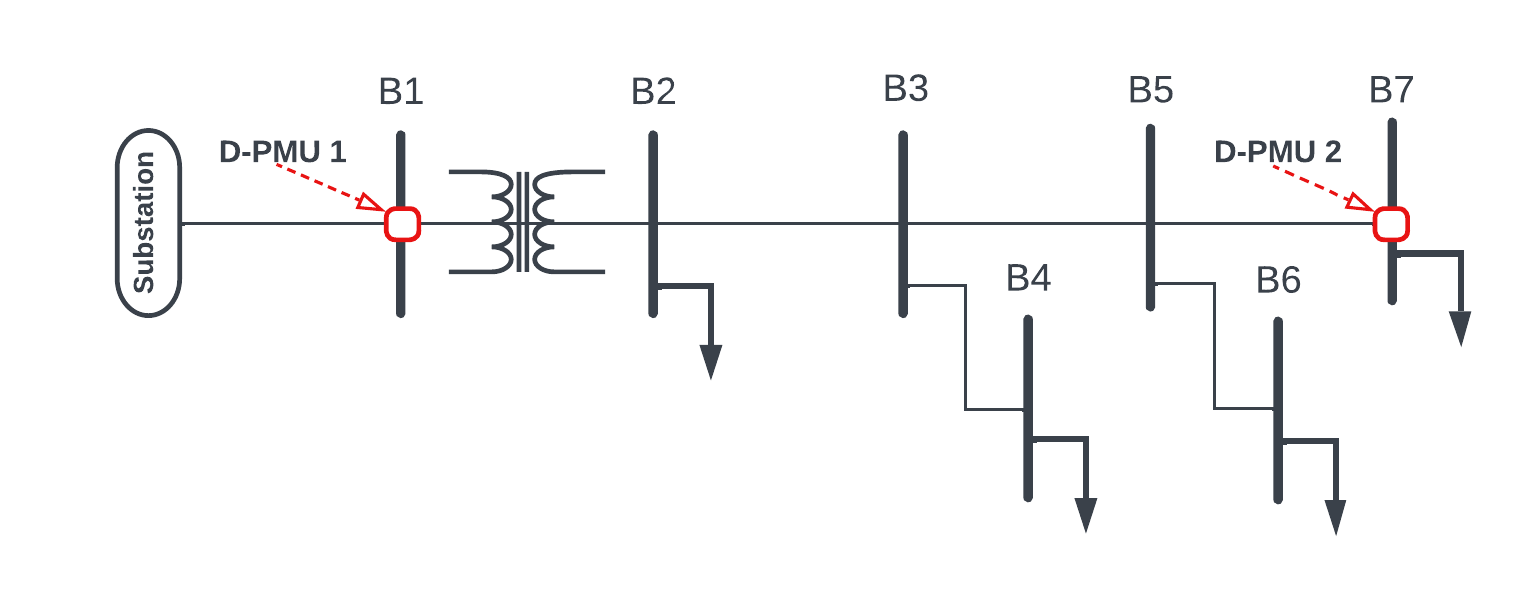}
\end{center}
\vspace{-0.35cm}
\caption{An example power distribution network with $N$ = 7 buses. Two PMUs are installed are installed at buses in B1 and B5. We have $\mathcal{M} = \{ B1, B7\}$.} 
\vspace{-0.2cm}\label{fig:7bus}

\end{figure}

Consider a power distribution system, such as the one in Fig. \ref{fig:7bus}. Let $\mathcal{B}$ denote the set of all buses, such as $\{B_1,\dots ,B_7\}$ in Fig. \ref{fig:7bus} and $N = \abs{\mathcal{B}}$ denotes the number of buses. Also let $\mathcal{M}$ denote the set of those buses that are equipped with D-PMUs, such as $\{B_1, B_7\}$ in Fig. \ref{fig:7bus}. For now, suppose the D-PMUs only provide the measurements for the fundamental phasors. The case where D-PMUs also act as H-PMUs to measure harmonic phasors will be discussed later in Section \ref{sec:Harmonic}.

When an event occurs, its impact is \emph{simultanously} captured by \emph{all} the D-PMUs at the buses in set $\mathcal{M}$. Let $X_i^j$ denote the \emph{time series} of the phasor measurements that are captured during event $i$ by the D-PMU at bus $j$, where $j \in \mathcal{M}$. Similar to \cite{arminjournal}, such time series is assumed to be a \emph{window} of the following measurements at a given D-PMU: the per-phase magnitude of voltage $V_\phi$, the per-phase magnitude of current $I_\phi$, and the per-phase power factor $\text{PF}_\phi$, where $\phi$ is the given phase, i.e., $\phi \in \{A, B, C\}$. The reason for using these measurements is to remove the impact of off-nominal frequencies in the phase angle measurements, see \cite[pp. 113]{Hamed_book}. 

As for the buses in set $\mathcal{B} \backslash \mathcal{M}$, we do \emph{not} have any phasor measurement available at these buses. Therefore, we inevitably  assume a \emph{constant} value, i.e., a flat time series, for $V_\phi$, $I_\phi$, and $\text{PF}_\phi$ at these buses during the event. We obtain such constants by running a simple steady-state power flow analysis based on the \emph{nominal load} (NL) at each bus. Such analysis is readily available in practice by using the utility's models of its feeders in standard software, such as CYME \cite{cyme} and Synergy \cite{synergi}.

\subsection{Graph Learning Approach} \label{subsec:approach}

In this paper, we use GNN to conduct \emph{topology-based representation learning}. GNN is the general framework for defining deep neural networks based on \emph{graph data} \cite{graphbook}. 

To benefit from the GNN attributes, we need to translate the power system topology and the measurements into \emph{graph-structured data}. Suppose $A$ is 
\emph{the adjacency matrix} for the graph of the power distribution network, where each node in the graph is a bus and each link in the graph is a distribution line. 
For each event $i$, we define an \emph{event graph}, denoted by $G_i$, which has the same adjacency matrix $A$. For each node $j$ in graph $G_i$, we define $X_i^j$ as the input matrix. In this regard, if we have the measurements for $M$ events in the data set, then the set of graph-structured data can be shown as:
\begin{equation}
         \{ G_1, G_2, \dots, G_M\}.
        \label{eq:gm}
\end{equation}

The main reason for using GNN is to encode the graph-structured data $G_i$ for each event $i$ to a single graph-level representation vector with low dimension, which incorporates both the measurements at event $i$ and the system topology. Such low-dimension representation helps to achieve a more accurate, interpretive and distinctive event clustering outcome. 

Similar to the neural networks (NNs), GNNs can include multiple hidden layers with trainable weights. However, GNNs also take into account the graph topology or the adjacency matrix. This means that, each nodal vector data at any hidden layer in a GNN is updated based on not only its own trainable weights, but also its \emph{neighboring nodes}' nodal vector data. 

To see the importance of differences and similarities between NNs and GNNs in the context of the analysis in this paper, let us define $X_{i}$ as the input matrix for graph $G_i$, such that row $j$ of matrix $X_i$ is the stacked vector of time series in $X_i^j$. Also, let us define $H_{i}^k$ as the hidden matrix data for graph $G_i$ at hidden layer $k$. We note that $H_{i}^0$ = $X_{i}$. In a common NN model we obtain the input matrix on the next layer by conducting forward propagation such as:
\begin{equation}
        \begin{multlined}
           H_{i}^{k+1} = \sigma(H_{i}^kW^k),
        \end{multlined}
        \label{eq:nn}
\end{equation}
where $\sigma$ is the activation function, e.g., ReLU($x$) = max$(0, x)$, and $W^k$ is the trainable weight matrix of layer $k$. 

In the NN framework, the input matrix $X_i$ and the hidden layer matrix $H_i^k$ contain different samples in their rows, which are often independent and identically distributed random variables. However, when it comes to a GNN, these samples (rows of data) are \emph{related to each other}. In this paper, these samples are the nodal data at each bus, which are simultaneously captured during the same event, but from the viewpoint of the sensors at different buses on the power distribution system. 
These samples are related to each other through the physics of the distribution system and the network topology. In the GNN framework, each nodal hidden layer data is updated based on its own NN output as well as its neighbours' NN output by using the adjacency matrix $A$. The revision of the equation in (\ref{eq:nn}) for the case of GNN will be given in Section II-B. 

In the next three sub-sections, we will explain how to implement our proposed graph learning method. In Section \ref{sec: graphencoder}, we will build an unsupervised GNN-based graph encoder to transform each $G_i$ to a single vector. In Section \ref{sec: mutual} we will set the objective of the graph encoder to maximize the mutual information between its node-level data and its graph-level data. Finally, in Section \ref{sec:Discriminator} we will develop a discriminator module to calculate the aforementioned mutual information.
\color{black}

\subsection{Graph Encoder}\label{sec: graphencoder}

In this section, we develop a GNN-based graph encoder, denoted by $\mathcal{E}$, in order to learn a \emph{single vector} that summarizes the time series for each graph-structured data $G_i$. Such vector will ultimately serve as the graph-level representation for each event. It is obtained by encoding the underlying shared properties of the data based on the topology of the system. The encoding process is based on maximizing mutual information between the node-level representation at each bus and the graph-level representation, which involves all of the buses. 


We construct the graph encoder by using GCN \cite{gcn} with the following 
updating formulation in its hidden layers\footnote{Other similar modules, such as those  in \cite{gat} and \cite{GraphSAGE}, can also be used.}:

\begin{equation}
        \begin{multlined}
           H_{i}^{k+1} = \sigma(D^{-\frac{1}{2}}\tilde{A}D^{-\frac{1}{2}}H_{i}^k\omega^k).
        \end{multlined}
        \label{eq:gcn}
\end{equation}

\noindent Here, $\tilde{A} = A + I_N$  is the adjacency matrix with added self-connections, $D$ is the \emph{degree matrix}, where $D_{aa} = \Sigma_{b}\tilde{A}_{ab}$, and $\omega^k$ is the set of trainable weights in the $k^{th}$ layer of the GNN.
The main difference between the formulation in (\ref{eq:gcn}) and the one in (\ref{eq:nn}) is the use of $D^{-\frac{1}{2}}\tilde{A}D^{-\frac{1}{2}}$, which aggregates the nodal data from neighbouring nodes. This element also symmetrically normalizes the rows of matrix $H_{i}^{k+1}$ cf. \cite{gcn}.

For each graph $G_i$ and each hidden layer $k$, let $\textbf{h}_i^{k}(j)$ denote row $j$ of matrix $H_i^k$. We refer to 
$\textbf{h}_i^{k}(j)$ as the node-level (or \emph{local}) representation of the event. Accordingly, for each node let us put together all such node-level representations at all the layers $k = 1, \ldots, K$ as follows:

\begin{equation}
        \begin{multlined}
           \textbf{h}_{i}^{\omega}(j) = [\textbf{h}_i^{1}(j), \textbf{h}_i^{2}(j), \dots, \textbf{h}_i^{K}(j)].
        \end{multlined}
        \label{eq:nodelevel}
\end{equation}

\noindent Superscript $\omega$ in $\textbf{h}_i^\omega(j)$ indicates the set of parameters for graph encoder $\mathcal{E}$. Furthermore, let us define: 
%
\begin{equation}
        \begin{multlined}
           \textbf{h}_{i}^{\omega, g} = S(\{\textbf{h}_{i}^{\omega}(1), \dots, \textbf{h}_{i}^{\omega}(N)\}) 
        \end{multlined}
        \label{eq:graphlevel}
\end{equation}
as the graph-level (or \emph{global}) representation of event $i$, where $S$ is a permutation invariant function that summarizes the node-level representation vectors to a single graph-level representation vector, such as via element-wise mean or max functions \cite{graphbook}. The  outputs of graph encoder are obtained as:
\begin{equation}
        \begin{multlined}
            \{\textbf{h}_{i}^{\omega}(j), \textbf{h}_{i}^{\omega, g}\} = \mathcal{E}(G_i) , \ \  \forall j = 1, \dots, N,
        \end{multlined}
        \label{eq:graph encoder}
\end{equation}

\noindent which include all the $N$ node-level representations and a single graph-level representation for each event $i$. The objective in the design of the graph encoder is to find the structural dependencies among the vectors from that are listed in (\ref{eq:graph encoder}).  

\subsection{Mutual Information}\label{sec: mutual}
Similar to the unsupervised learning methods  in \cite{DIM} and \cite{DGI}, the objective for the proposed graph encoder is 
to maximize the average mutual information (MI) \cite{MI} of the graph-level representation $\textbf{h}_{i}^{\omega, g}$ and all of the node-level representations $\textbf{h}_{i}^{\omega}(j)$ for any $j=1, \dots, N$ for each event $i$ as:
\begin{equation}
        \begin{multlined}
            {\text{Maximize:} \: \; \mathcal{I}=\frac{1}{M} {\sum_{j=1}^{M} \frac{1}{N} \sum_{i=1}^{N} \text{MI}(\textbf{h}_{i}^{\omega}(j); \textbf{h}_{i}^{\omega, g})}}.
        \end{multlined}
        \label{eq:mi_all_events}
\end{equation}
The above maximization enforces the GNN graph-level representation to carry the type of information that is present in all of the nodes in the network and all the layers \cite{DGI}. 
It should be mentioned that, in this paper, we focus on graph-level representation learning, rather than on substructure representation learning. The latter strictly focuses on node-level tasks, such as for the node classification in \cite{gcn}. 

Calculating $\mathcal{I}$, in a continuous and high-dimensional settings is difficult. A solution is suggested in \cite{mine}, in which we use a mutual information \emph{estimator} between the input and the output of the deep neural networks. This method, is based on training a classifier (a discriminator) that separates samples from the joint distribution and their product of marginals.


\color{black}

\vspace{-0.1 cm}

\subsection{Discriminator: Positive and Negative Graphs. }\label{sec:Discriminator}

The first step in \emph{estimating} the mutual information is 
to define the joint and marginal distributions.
The joint distribution, i.e., positive samples in this paper, are defined as node-level/graph-level representation pairs ($\textbf{h}_{i}^{\omega}(j), \textbf{h}_{i}^{\omega, g}$), for each \emph{actual event} $G_i$. Also, we refer to $G_i$ as \emph{positive graphs}.

We also need to construct negative samples  in order to define the product of marginals. Note that, the choice of the negative samples has impact on the type of structural information that is desirable to be captured as a byproduct of estimating MI \cite{DGI}. First, we construct the \emph{negative graphs}. They have the same input node data as in the positive graphs, i.e., $X_i^j$ for event $i$ and node $j$. But they have a different 
graph topology, which are \emph{random trees} with the same number of nodes and links.
Next, we consider any pair of a graph-level representation from an actual event graph (a positive graph) with any node-level representation of a negative graph, that are obtained from graph encoder, as a negative sample.

\begin{figure}[t]
\begin{center}
\includegraphics[trim={0.5cm 0.5cm 0.5cm 0.5cm},clip,scale=0.75]{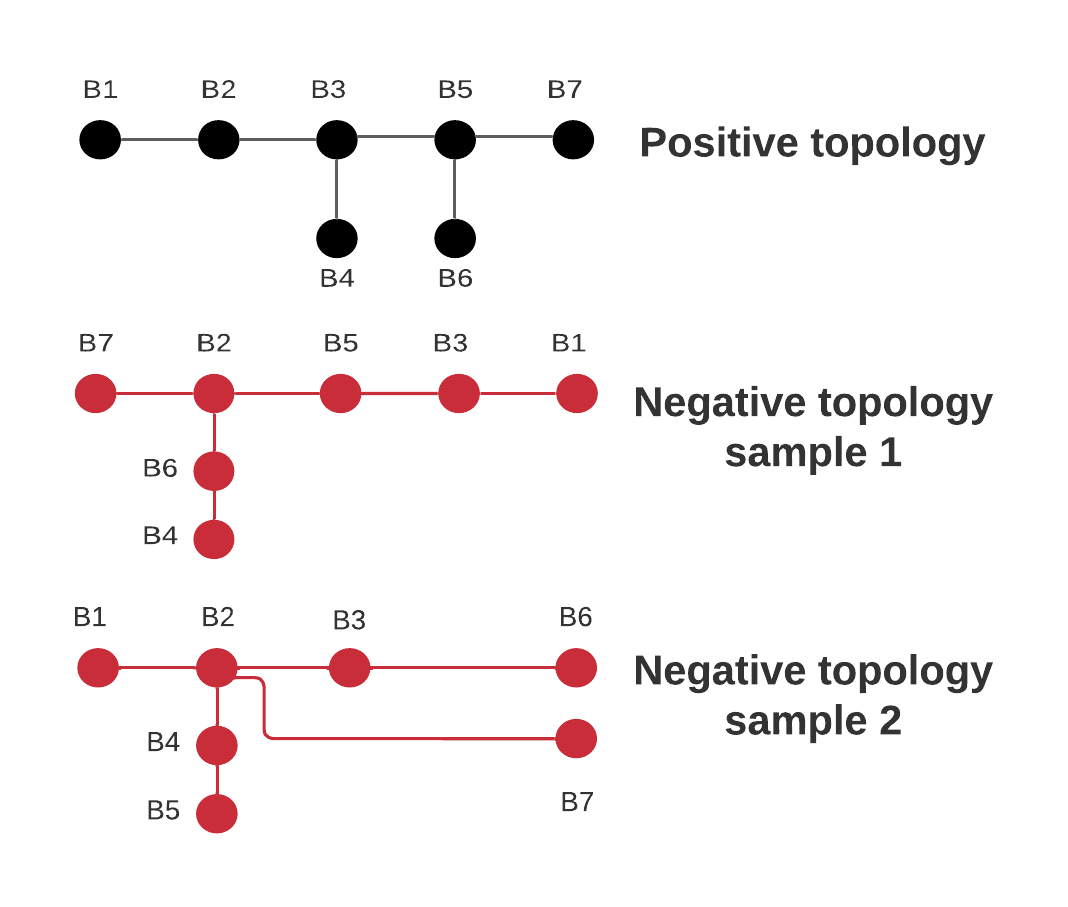}
\end{center}
\vspace{-0.3cm}
\caption{The actual (i.e., \emph{positive}) topology of the power distribution network in Fig. 1 is shown in black. Two arbitrary alternative  (i.e., \emph{negative}) topologies with the \emph{same} number of nodes/buses and  edges/lines are shown in red.
%
} \label{fig:posneggraphs}
\vspace{-0.3cm}
\end{figure}

The concept of positive graphs and negative graphs is illustrated in an example in Fig. \ref{fig:posneggraphs}. 
The positive topology represents the actual topology of the power distribution system that we saw in Fig. \ref{fig:7bus}. The other two arbitrary negative topologies are used to shape two samples of negative graphs.
These positive and negative samples are used to train the discriminator-based method that is proposed in \cite{mine}. This approach enforces the encoder to learn the structural dependency of the data, which leads to an overall MI maximization in an average sense.


In this study, discriminator $\Psi$ is a neural network with a set of parameters $\psi$. We set the discriminator to output 1 for a positive sample and a 0 for negative sample. 
%
%
%
%
%
Accordingly, based on the Jensen-Shannon MI estimator that is proposed in \cite{DIM}, and the method in \cite{mine}, we \emph{simultaneously estimate and maximize} the objective function $\mathcal{I}$ in (\ref{eq:mi_all_events}) as shown below:
\begin{equation}
        \begin{multlined}
            \hat{\mathcal{I}}_{\omega,\psi} = 
            \frac{1}{2MN}\sum_{i=1}^{M}\Big(\sum_{j=1}^{N}
            \mathbb{E}_{\mathbb{P}}[-\sigma(
            -\Psi(\textbf{h}_{i}^{\omega}(j), \textbf{h}_{i}^{\omega, g})
            ]\\
            -\sum_{j=1}^{N}\mathbb{E}_{\mathbb{P} \times \mathbb{P^{'}}}[\sigma(
            \Psi(\textbf{h}_{i^{'}}^{\omega}(j), \textbf{h}_{i}^{\omega, g})
            ]\Big),
        \end{multlined}
        \label{eq:JSD_MI}
\end{equation}
%
%

%
%
\noindent 
where $\hat{\mathcal{I}}_{\omega,\psi}$ is the Jensen-Shannon MI estimator; and $\sigma(x) = log(1+e^x)$ is the softplus function. In this study, for each physical (i.e., positive) event $i$, we make a corresponding random negative event $i^{'}$. Accordingly, the probability distribution of the positive events, which is denoted by $\mathbb{P}$, is identical to the probability distribution of the negative events, which is denoted by $\mathbb{P^{'}}$. Also, $\mathbb{E}_{\mathbb{P}}$ and $\mathbb{E}_{\mathbb{P} \times \mathbb{P^{'}}}$ are the expected value for the discriminator output related to the positive samples (or the joint distribution) and the negative samples (or the product of marginals), respectively.
Due to the summation, for both negative and positive samples, we include the coefficient $1/2$ in (\ref{eq:JSD_MI}).  
Notations $\textbf{h}_{i}^{\omega}(j), \textbf{h}_{i^{'}}^{\omega}(j)$ and  $\textbf{h}_{i}^{\omega, g}$ indicate the outputs of the graph encoder $\mathcal{E}$, and represent the node-level representation of the positive event $i$ in node $j$, the node-level representation of the negative event $i^{'}$ in node $j$, and the graph-level representation of the positive event $i$, respectively. 

\begin{algorithm}[t]

    \caption{Topology-Based Representation Learning}
  \begin{algorithmic}[1]
     \label{alg:sp}

      \STATE \textbf{Input:} Event time series $X_i^j$ and network topology $A$.

      \STATE \textbf{Output:} Graph-level representation vectors clusters.

     \STATE \textbf{// Positive and Negative Graphs}

     \STATE \textbf{For} each training event $i$ \textbf{Do}
     \STATE $ \ \ \, $Construct the positive graphs and assign $X_i^j$ to all nodes.
     \STATE $ \ \ \, $Construct the negative graphs and assign $X_i^j$ to all nodes.
     \STATE \textbf{End}
    
    
    \STATE \textbf{// Training Graph Encoder and Discriminator}
    \STATE \textbf{For} each epoch of training data \textbf{Do}
    \STATE $ \ \ $ Obtain $\{\textbf{h}_{g_b}^{\omega}(j)$, $\textbf{h}_{g_b}^{\omega, g}\} = \mathcal{E}(g_b)$ for all positive graphs.  
    
    \STATE $ \ \ $ Obtain $\{\textbf{h}_{g'_b}^{\omega}(j)$\}$ = \mathcal{E}(g'_b)$ for all negative graphs.

    \STATE $ \ \ $ Pair each $\textbf{h}_{g_b}^{\omega}(j)$ with its relative $\textbf{h}_{g_b}^{\omega, g}$ as positive
    
    $ \ \ $ sample; and label the discriminator's output as 1.
    
    \STATE $ \ \ $ Pair each $\textbf{h}_{g'_b}^{\omega}(j)$ with its relative $\textbf{h}_{g_b}^{\omega, g}$ as negative 
    
    $ \ \ $ sample; and label the discriminator's output as 0.
    
    
    \STATE $ \ \ $ Calculate the loss function in (\ref{eq:JSD_MI}).
    \STATE $ \ \ $ Update the $\omega$ and $\psi$ by conducting back propagation 
    
    $ \ \ $ and using Adam optimizer \cite{adam}.
    \STATE \textbf{End}

    \STATE \textbf{// Graph-level Representation}
     \STATE \textbf{For} each graph $G_i$ \textbf{Do}
    \STATE $ \ \ \  \ \ \ $ Obtain \{$\textbf{h}_{i}^{\omega, g}\} = \mathcal{E}(G_i)$.
    \STATE \textbf{End}
    \STATE \textbf{// Clustering}
    \STATE Cluster the event vectors $\{\textbf{h}_{i}^{\omega, g}\}$ using GMM.
  \end{algorithmic}
\end{algorithm}

\vspace{-0.2cm}
\subsection{Clustering} \label{subsec:clustering}
After training the graph encoder, the graph-level representations $\textbf{h}_{i}^{\omega, g}$, are obtained for all events $i=1,\dots,M$, and they are clustered by using the Gaussian Mixture Model (GMM). The GMM uses expectation maximization algorithm for fitting a mixture of Gaussian models to the training data set, considering a pre-defined number of clusters. Then, each $\textbf{h}_{i}^{\omega, g}$ is assigned to the most probable cluster.

Note that, the purpose of our proposed method is to properly incorporate the topological information from the sensor measurements to learn the most distinctive representation for the type of each event, such that we can enhance the clustering accuracy with the already existing clustering methods. 

We shall note that, we did examine other clustering methods, such as K-means and DBSCAN; however, GMM demonstrated the highest average clustering performance.

\vspace{0.05cm}
\subsection{Algorithm: Topology-Based Representation Learning}

Algorithm \ref{alg:sp} shows the summary of the steps that we took in Sections \ref{sec: graphencoder} to \ref{subsec:clustering}. It is divided into four segments. First, we generate the positive and negative graphs; see lines 4 to 7. Next, we train the graph encoder and the discriminator; see lines 9 to 16. Third, we obtain the graph-level representations for all the events; see lines 18 to 20. Finally, we do the clustering task using GMM and based on the obtained graph-level representations of the events; see line 22.


\section{Temporal Representation Learning} \label{sec:temporal}
The design that we presented in Section \ref{sec:topology} can fully incorporate the knowledge about the topology of the network and the  relative location of the measurements into the task of event clustering. However, if we use the method in Section \ref{sec:topology} \emph{as is}, then it may \emph{not} result in a significant improvement compared to some benchmark methods in the literature. The main issue here is the \emph{high dimentionality} in the time series that needs to be placed at each node of the graph in this field. 

\subsection{Tackling High Dimensionality}
To address the above issue, we propose to compress the data in the time series by learning the temporal-dependent features of the events. 
By compressing the event data in time domain, we can lower the dimension of the feature space. This leads to achieving a higher computational and clustering efficiency with less numerical challenges.




Accordingly, an Auto-Encoder-Decoder (AED) \cite{aed} model is proposed which includes Long Short Term Modules (LSTM) \cite{lstm} for proper temporal-based representation learning of each node time series for each event.
AED constitutes of two main parts. The first part is the temporal encoder ($E$), which tries to summarize and transfer each event time series matrix $X_i^j$ into a 
single embedding vector (${Em_i^j}$). The second part is the temporal decoder ($D$), which tries to reconstructs the actual time series with the mentioned embedding vectors.

The objective function of AED is to minimize the Mean Square Error (MSE) of the time series input to $E$ and the  time series output of $D$. Here are the details of the AED model: 
\begin{equation}
        \begin{multlined}
           Em_i^j = E(X_i^j), \forall i=1,..., M, \forall j=1, ..., N,
        \end{multlined}
        \label{eq:aedEncoder}
\end{equation}
\begin{equation}
        \begin{multlined}
           \theta_E, \theta_D = \underset{\theta_E, \theta_D}{\mathrm{argmin}} \, \frac{1}{M} \sum_{j=1}^{M} \frac{1}{N} \sum_{i=1}^{N}\text{MSE}(X_i^j, D({Em_i^j})),
        \end{multlined}
        \label{eq:mse}
\end{equation}
which $\theta_E, \theta_D$ are the encoder deep neural network and decoder deep neural network parameters, respectively. 

\begin{figure*}[b]
\begin{center}
\includegraphics[trim={0.5cm 0.5cm 0.5cm 0.5cm},clip,scale=0.52]{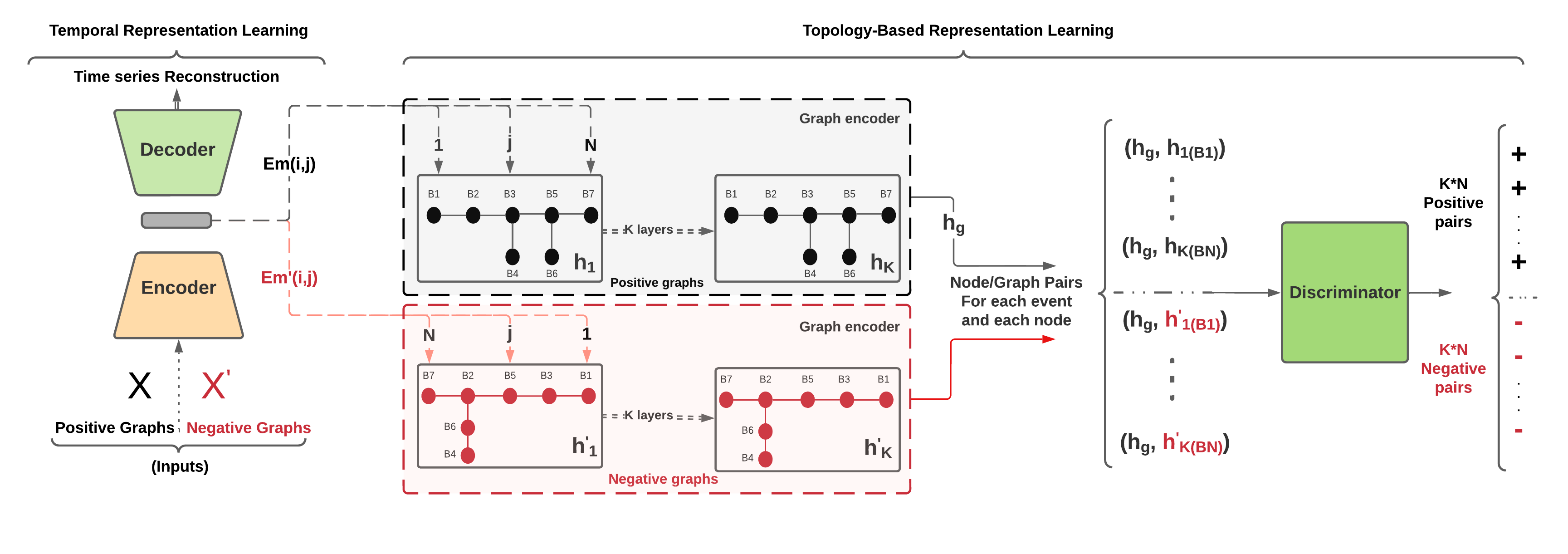}
\end{center}
\vspace{-0.65cm}
\caption{GraphPMU: AED learns the optimal representation vectors for nodal data for all events. Then these vectors alongside the positive and negative graphs are used as input for the proposed graph encoder. The output of the graph encoder, node-level and graph-level representations are paired to shape the positive and negative samples. Then, discriminator learns to discriminate between these samples for MI maximization. 
} \label{fig:gpmuprocess}
\vspace{-0.3cm}
\end{figure*}

In this study, the parameters of AED are shared between all the buses. This means that during the training phase, for any event in any bus, the AED parameters are being updated. In other words, instead of considering multiple AEDs for each bus, we rather implement a global AED. This makes the training process faster. It also allows the AED model to take advantage of the learned features from different  locational time series. This prevents an over-fitting over a single bus data. After training the AED model, all event time series data for each node $X_i^j$ can be encoded to their $Em_i^j$ by using (\ref{eq:aedEncoder}). 

\subsection{Algorithm: Temporal Representation Learning}
Algorithm \ref{alg:temp} shows the steps for temporal representation learning. 
First, we train the temporal encoder and decoder; see lines 4 to 8. After that, we obtain the compressed embedding vector for all events and nodes time series; see lines 10 to 12.

\begin{algorithm}[t]

    \caption{ Temporal Representation Learning}
  \begin{algorithmic}[1]
     \label{alg:temp}
      \STATE \textbf{Input:} Normalized event time series $X_i^j$. 
    
      \STATE \textbf{Output:} Embedding vectors $Em_i^j$ as in (\ref{eq:aedEncoder}).

    \STATE \textbf{// Training Phase}
    \STATE \textbf{For} each epoch \textbf{Do}

    \STATE $ \ \ \ $ Pass $X_b$ to the temporal AED ($E$ and $D$).
    \STATE $ \ \ \ $ Calculate loss function from (\ref{eq:mse}).
    \STATE $ \ \ \ $ Update $\theta_E$ and $\theta_D$ through back propagation \cite{adam}.
    
    \STATE \textbf{End}

    \STATE \textbf{// Embedding Extraction}
     \STATE \textbf{For} each event $i$ and node $j$ \textbf{Do}
    \STATE $ \ \ \ $ $Em_i^j = E(X_i^j)$.
    \STATE \textbf{End}
  \end{algorithmic}
\end{algorithm}


\vspace{0.05cm}
\section{GraphPMU: Combining Topology-Based and Temporal Representations}\label{sec:GraphPMU}

We are now ready to introduce our ultimate GraphPMU method by combining the topology-based representation learning design in Section \ref{sec:topology} with the temporal representation learning design in Section \ref{sec:temporal}. Fig. \ref{fig:gpmuprocess} shows how these two design components are integrated in order to achieve GraphPMU. 

The architecture in Fig. \ref{fig:gpmuprocess} can be explained by going through its parts from left to right. The process starts with training the time domain AED with matrices $X_i^j$. This step is independent from the network topology; hence, it is the same for positive and negative graphs, as they have the same input time series. 

After the AED is trained, the obtained embedding vectors $Em_i^j$ from the temporal encoder $E$, are used to train the GNN model. Subsequently, the positive and negative graphs are shaped based on the embedding vectors and the defined topologies in Section \ref{sec:Discriminator}. These positive and negative graphs are passed to the graph encoder $\mathcal{E}$ to construct the node-level and graph-level representations. Then the obtained positive and negative samples are used to train the discriminator $\psi$.



Given the models for ($E$ and $D$) and ($\mathcal{E}$ and $\Psi$), the graph-level representation vectors are obtained as $\textbf{h}_{i}^{\omega, g} = \mathcal{E}(E(X_i^j))$ for all buses $j \in \mathcal{B}$. These graph-level representations are then clustered by the GMM method as we explained in Section \ref{subsec:clustering}.



\section{Extension to Incorporate Harmonic Synchro-Phasors} \label{sec:Harmonic}
So far, we have assumed that all the phasor measurements are obtained at the fundamental frequency. This is indeed the state of practice in this field for a typical PMU. However, as we mentioned in Section \ref{sec:Introduction}, 
it is envisioned that standard D-PMUs may in the future also act as H-PMUs to provide the phasor measurements not only at the fundamental frequency but also at selected harmonic frequencies. Accordingly, in this section, we will expand the GraphPMU model to incorporate such emerging advancement in data availability in this field.

\subsection{More Distinctive Event Signatures} \label{subsec:har_data}
Without loss of generality, we assume that each D-PMU provides the synchronized phasor measurements for the 3rd and 5th harmonics, in addition to the fundamental frequency. Expanding the analysis to include higher harmonic orders would be similar, although it may not be necessary; because most events manifest themselves properly in either the 3rd or the 5th harmonics, or in both.
At each bus $j$ in $\mathcal{M}$, we collect $V_\phi$, $I_\phi$, and $\text{PF}_\phi$; however, this is done not only for the fundamental frequency but also for the 3rd and the 5th harmonics.  
In this regard, taking into account the harmonic phasors can be highly beneficial as they can demonstrate \emph{more distinctive signatures} for the purpose of clustering the events. This can help compensate for some of the challenges in having locationally-scarce measurements; thus, contributing to the overall success in the proposed GraphPMU method.

\begin{figure}[t]
\begin{center}
\includegraphics[trim={2cm 0.5cm 2cm 1.5cm},clip,scale=0.29]{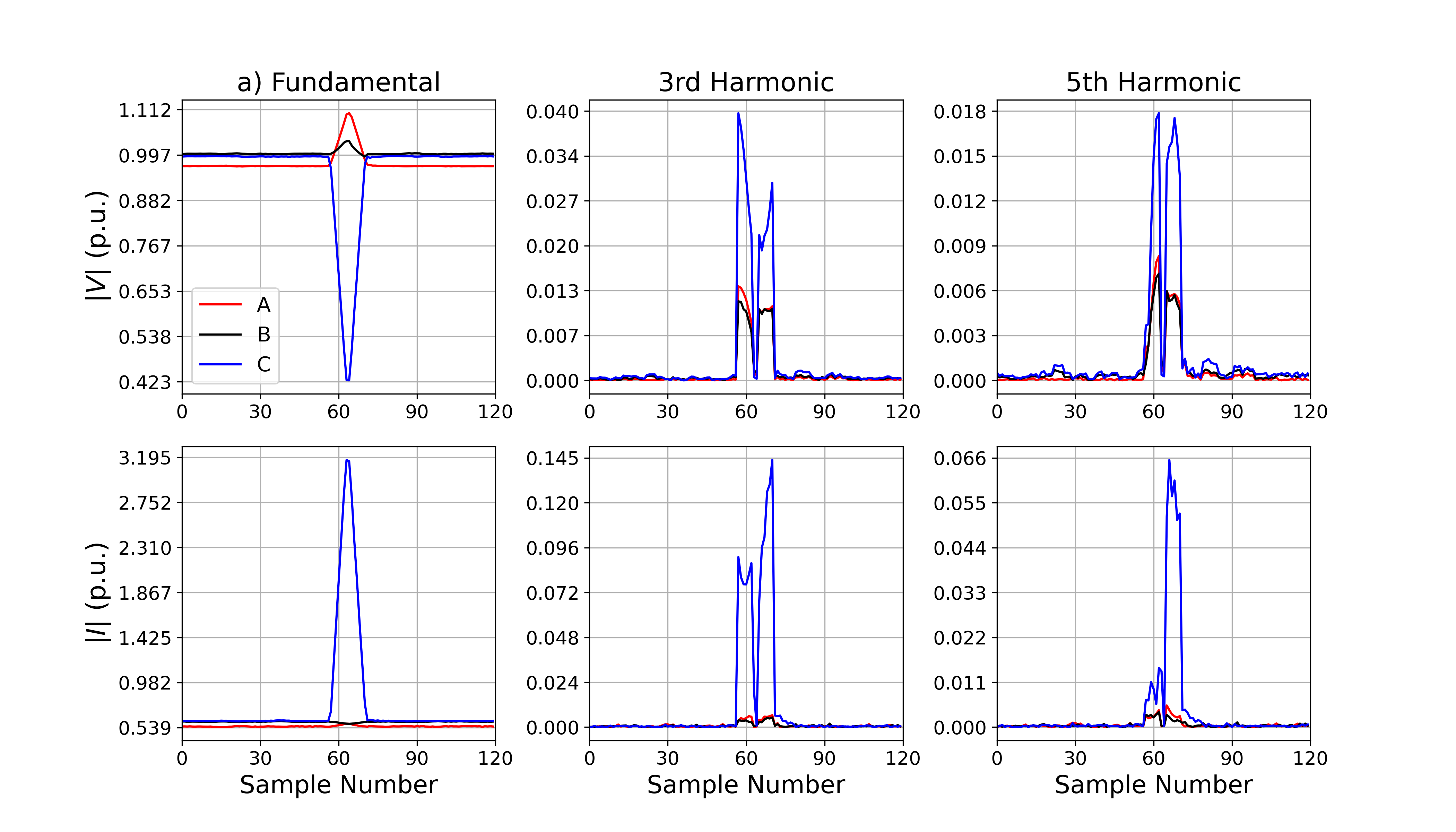}
\end{center}
\vspace{-0.35cm}
\caption{Comparing the signatures during the \emph{same} event at the fundamental phasor measurements vs. at the 3rd and 5th harmonic phasor measurements.} \label{fig:harm}
\vspace{-0.3cm}
\end{figure}

As an example, 
Fig. \ref{fig:harm} 
shows the event signatures in different types of phasor measurements during a single-line-to-ground fault. The event signature in the fundamental frequency in Fig. \ref{fig:harm}(a) 
is  a simple voltage sag and a simple inrush current. However, the event signatures in the harmonic phasor measurements at the 3rd harmonic in Fig. \ref{fig:harm}(b) and at the 5th harmonic in Fig. \ref{fig:harm}(c) are considerably more distinctive.

\subsection{Extended Temporal-Based Learning} \label{subsec:har_feat_extract}
Same as in the case for the fundamental phasors data, we use AED to learn time domain representations for the time series of the harmonic phasor measurements. Accordingly, we obtain the embedding vectors to use them in the clustering process. Importantly, since the strength and the overall nature of the time series of the harmonic phasors are different from those of the fundamental phasors,  we must train \emph{different} AEDs for \emph{each} fundamental or harmonic order. This makes the time domain representation learning more reliable and more accurate than using a single AED for these different time series. The training is done by using Algorithm \ref{alg:temp}. 

%
We concatenate all the embedding vectors to form:
\begin{equation}
        \begin{multlined}
           Em_i^{j} = [{Em_i^{j,1} \ Em_i^{j,3} \ Em_i^{j,5}}]^T.
        \end{multlined}
        \label{eq:harms}
\end{equation}


\subsection{Extended Topology-Based Learning}

Next, we feed the new vectors $Em_i^{j}$ that are derived in (\ref{eq:harms}) as the input to the GNN using Algorithm \ref{alg:sp} to complete the process for the event clustering task. Since the size and nature of the input vector is different from those in Section \ref{sec:topology}. We need to re-train the GraphPMU based on the new vector of features. Last but not least, for each bus $j \in \mathcal{B} \backslash \mathcal{M}$, which does not have a sensor, we use zero padding concatenation to the fundamental embedding vectors that the previously obtained in section \ref{sec:temporal}). This is because the default steady state values in unobserved locations are assumed not to have any harmonics.

\section{Case Studies} \label{sec:Results}

\begin{figure}[t]
\begin{center}
\vspace{-0.15cm}
\includegraphics[scale=0.34]{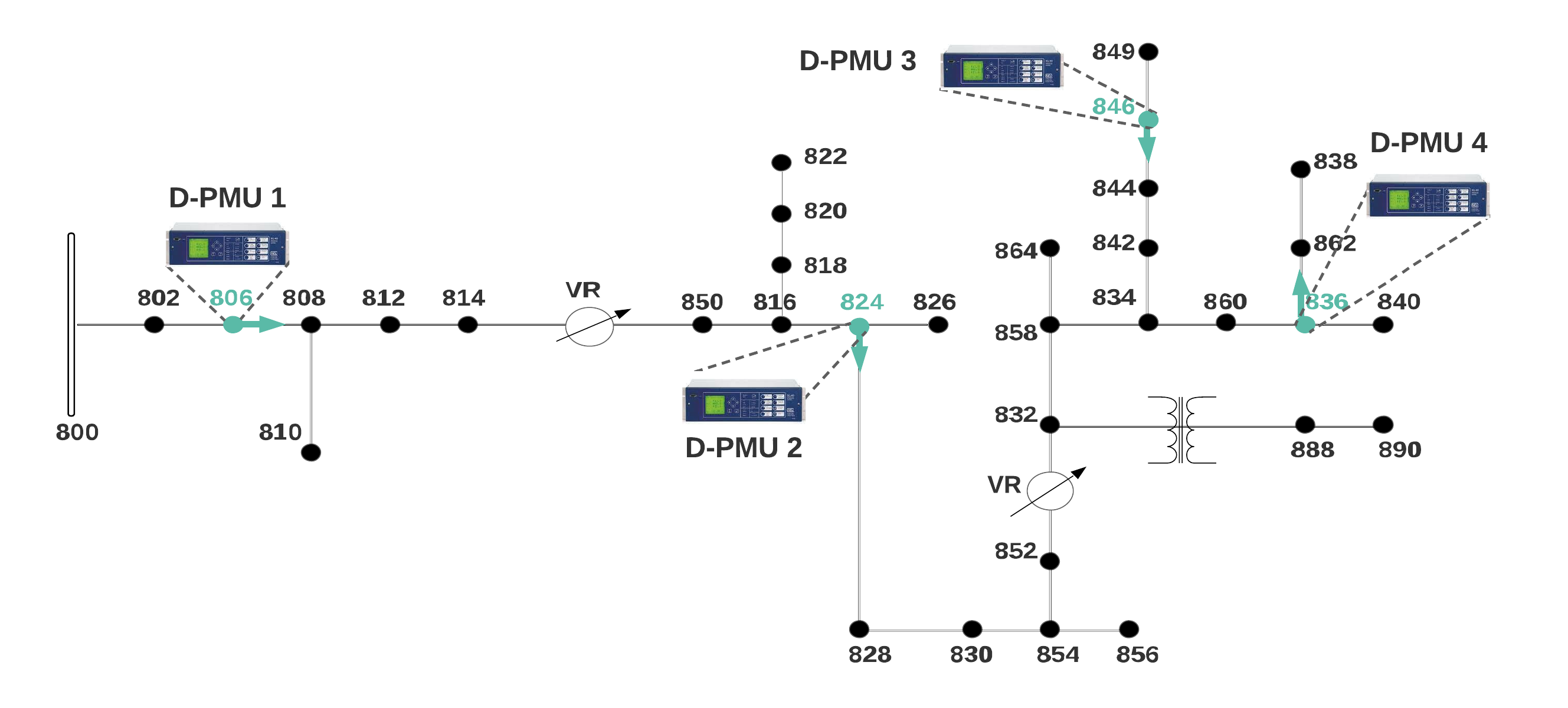}
\end{center}
\vspace{-0.45cm}
\caption{The IEEE 34-bus test system with locationally-scarce phasor measurements. There are only four D-PMUs (H-PMUs) available on this network, as marked on the figure. However, the events can happen at \emph{any} location.} \label{fig:IEEE}
\vspace{-0.4cm}
\end{figure}

In this section, we conduct various case studies based on the IEEE 34-bus three-phase power distribution test system, which is shown in Fig. \ref{fig:IEEE}. The network simulation model is developed in PSCAD to assure capturing the transient signatures of the events \cite{PSCAD_Reference}. Nine different types of events are simulated: 
\begin{enumerate}
\item Three-Phase Capacitor Bank switching at bus 840
\item Three-Phase Capacitor Bank switching at bus 849
\item Single-phase load switching at bus 858
\item Three-phase load switching at bus 836
\item Three-phase motor-load switching at bus 812
\item Three-phase motor load switching at bus 828
\item Single-phase-to-ground fault at bus 852
\item Two-phase-to-ground fault at bus 862
\item Three-phase-to-ground fault at bus 816.
\end{enumerate}

Unless stated otherwise, we assume that there are only four phasor measurement units are available on the power distribution network. The location of the D-PMUs (H-PMUs) are shown on Fig. \ref{fig:IEEE}. Note that, we have:
\begin{equation}
\mathcal{M} = \{806, 824, 836, 846\}. \label{eq:pmus}
\end{equation}
Depending on the case study, we assume that each D-PMU either provides the phasor measurements only for the fundamental component, or for the fundamental component together with the 3rd and the 5th harmonics. 
As in practice, we assume that events occur rarely \cite{arminjournal}; therefore, we assume only a small number of each type of event are available to train GraphPMU.  
We augmented the data from the few available events by conducting time shifting and adding noises to the raw data. This is done for each type of event and for all sensors. 
In total, we considered 50,000 events of various types for training, 5000 events for evaluation, and 5000 events for testing. 

\subsection{Parameters of GraphPMU} \label{sec: Parameters Detail}
The graph encoder has two layers of GCN \cite{gcn}, where the sizes of the vectors for the hidden layers are 128 and 64, respectively. The discriminator contains two fully-connected layers with 192 and 32 neurons. We concatenate the hidden layer features and the global graph features together; thus, the input size of discriminator is 128+64 = 192. By intentionally choosing a naive discriminator with only two fully-connected NNs, 
we enforce the GNN encoder to learn more discriminative features. This  can help with event clustering.

The encoder portion of the AED has two layers of LSTM modules with 32, and 64 units, following with a 32 units fully- connected layer. 
The decoder portion is an almost reverse version of the encoder, with a fully-connected $64\times125$ layer, followed by two LSTM layers with 64 and 32 units. 

All activation functions are LeakyReLU, 
where the slope of the leak is 0.2. For tuning the hyper-parameters, we used the \emph{coarse-to-fine} method \cite{arminjournal}. The learning rate $\alpha$ is $1e^{-3}$ for Adam optimizer, and $\beta_1$ is 0.5 for better stability in training. All models  are  developed with  Pytorch. The GNN models are built with the Deep Graph Library \cite{dgl_package}, by  using Nvidia  GTX  1050  ti  GPU  and  a  core  i-7  2.2GHz  CPU  with 32 GB RAM.

The MSE for the training and testing stages in the fundamental phasor are 0.04425 and 0.04522, respectively. This shows 
that the encoder is able to compress high-resolution data to a low dimension such that the decoder can reconstruct the time series with high accuracy. 
This confirms the performance of the AED sub-system for the rest of our analysis. 
In this paper, Adjusted Rand Index (ARI) score \cite{ARIref} is used to assess accuracy in a clustering task. ARI is a number between 0 and 1. A higher ARI means a better clustering. 

\subsection{Comparison with Temporal-Based Benchmarks} \label{subsec: time series clustering result}

\begin{table}[t]


\centering
    \caption{ARI Score for Different Methods Under \\ Locationally-Scare Phasor Measurements at Four Buses}
    \label{table:clustering}

    \scalebox{1.1}{
        \begin{tabular}{|c|c|c|}
            \cline{2-3}
            \multicolumn{1}{c|}{} & \multicolumn{1}{c|}{Method} & \multicolumn{1}{c|}{ARI} \\\cline{1-3} 
            \parbox[t]{0.4cm}{\multirow{5}{*}{\rotatebox[origin=c]{90}{Without} \rotatebox[origin=c]{90}{Graph Model}}} & \multicolumn{1}{c|}{AED} & \multicolumn{1}{c|}{0.473} \\\cline{2-3}
            & \multicolumn{1}{c|}{DEC} & \multicolumn{1}{c|}{0.520}  \\\cline{2-3}
            & \multicolumn{1}{c|}{Kernel TS} & \multicolumn{1}{c|}{0.237}  \\\cline{2-3}
            & \multicolumn{1}{c|}{$k$-shape TS} & \multicolumn{1}{c|}{0.418}  \\\cline{2-3}
                                     & \multicolumn{1}{c|}{$k$-means TS} & \multicolumn{1}{c|}{0.343} \\\hline \hline
                                 
             \parbox[t]{0.4cm}{\multirow{5}{*}{\rotatebox[origin=c]{90}{With} \rotatebox[origin=c]{90}{Graph Model}}} & \multicolumn{1}{c|}{TS + N/G + NL} & \multicolumn{1}{c|}{0.487} \\\cline{2-3}
                                 & \multicolumn{1}{c|}{AED + N/G } & \multicolumn{1}{c|}{0.423} \\\cline{2-3}
                                 & \multicolumn{1}{c|}{AED + N/G + RL} & \multicolumn{1}{c|}{0.533} \\\cline{2-3}
                                 & \multicolumn{1}{c|}{AED + G + NL} & \multicolumn{1}{c|}{0.585} \\\cline{2-3}
                                 & \multicolumn{1}{c|}{\textbf{GraphPMU = AED + N/G + NL}} & \multicolumn{1}{c|}{{\textbf{0.720}}} \\\hline

        \end{tabular}
        }
     \vspace{ -0.05cm}
    \label{tab:models_ARI}


\end{table}

\normalsize

Table \ref{tab:models_ARI} shows the ARI for the proposed event clustering method (in the last row), in comparison with several benchmark methods (in the first nine rows). 
In this section, our focus is on the \emph{top segment} in Table \ref{tab:models_ARI}, i.e., the first five methods. These are the methods that do \emph{not} use any information about the network topology. These five methods are AED \cite{aed}, DEC \cite{dec},  Kernel $k$-means, $k$-Shape clustering and $k$-means clustering methods \cite{tslearn}.
Here, Time Series (TS) means that the method uses the raw time series data, without any encoding. 

From Table \ref{tab:models_ARI}, among the five methods that do \emph{not} use graph models, DEC and AED have the highest accuracy. 
To have a fair comparison, we assume the same steady-state constants at the buses without sensors for all the ten methods in Table \ref{tab:models_ARI}. 

\subsection{Comparison with Topology-Based Benchmarks} \label{subsec: graph based clustering result}

The next five methods in the \emph{bottom segment} of Table \ref{tab:models_ARI} \emph{do} use the information about the network topology. All of these combinations could have been used for our purpose. However, only the last row shows our ultimate design for GraphPMU. The rest of the methods serve as benchmarks. Regarding the new abbreviations in Table \ref{tab:models_ARI}, G means using only the graph-level representation in the GNN, N/G means using both the node-level and the graph-level representations, NL means using the nominal load flow model to obtain the constants at the buses with no sensors, RL means using random loading data instead of using nominal loading data.

\vspace{0.05cm}
\subsubsection{Advantage of Using Data Compression}
If we compare TS+N/G+NL versus  GraphPMU  in Table \ref{tab:models_ARI}, we can see that their difference is only in the use of AED instead of TS. Importantly, since the input to the GNN is more compressed in  GraphPMU, it becomes more distinctive for the GNN, as opposed to using the raw time series in TS+N/G+NL. Thus, the overall performance in event clustering is much better for the GraphPMU. Nevertheless, the use of topology information in TS+N/G+NL can still outperform most of the benchmark methods in the top segment of Table \ref{tab:models_ARI} that do \emph{not} use any graph model. 

\vspace{0.05cm}
\subsubsection{Advantage of Pairing Node-Level and Graph-Level Vectors}
If we compare AED+G+NL versus  GraphPMU  in Table \ref{tab:models_ARI},  we can see that their difference is only in terms of using G versus N/G. The method with AED+G+LN considers only the last layer of the graph learning model for the positive and the negative graphs for the MI maximization, rather than using the node-level/graph-level pairs. However, the use of such pair in GraphPMU is necessary to properly extract the shared structure between the node-level  and the graph-level representations, in order to have more distinctive clusters.

\vspace{0.05cm}

\subsubsection{Advantage of Using Nominal Load Data}
If we compare AED+N/G versus GraphPMU  in Table \ref{tab:models_ARI},  we can see that their difference is only in terms of using N/L. In AED+N/G, we do \emph{not} include the buses with no sensors in  graph-based learning. As a result, the accuracy of the method drops significantly. The reason is that there are only four nodes on the graph, i.e., the four buses with sensors. This is due to the locational scarcity of the sensors. Such a  small graph does \emph{not} give much room to benefit from topology-based learning. As for AED+N/G+RL, this method too suffers a considerable drop in performance. 
These results confirm that we \emph{do} benefit from conducting a simple power flow analysis based on the nominal loading data.

\begin{figure}
\centering
    \subfloat[DEC, ARI = 0.524]{\label{sublable1}\includegraphics[trim={3.1cm 1cm 4cm 3cm},clip,scale=0.2]{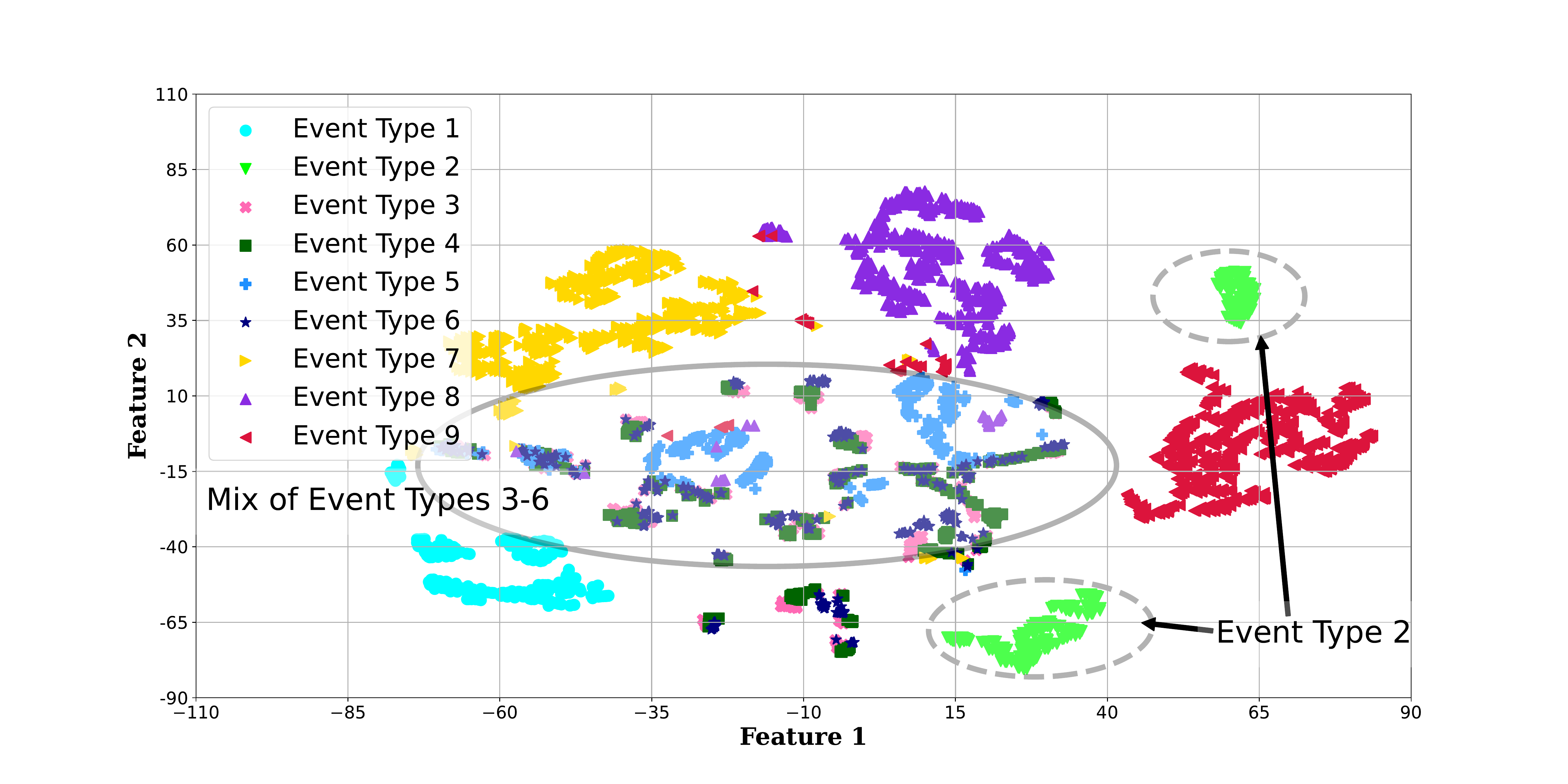}}
    
    \subfloat[AED+G+NL, ARI = 0.585]{\label{sublable2}\includegraphics[trim={3.1cm 1cm 4cm 3cm},clip,scale=0.2]{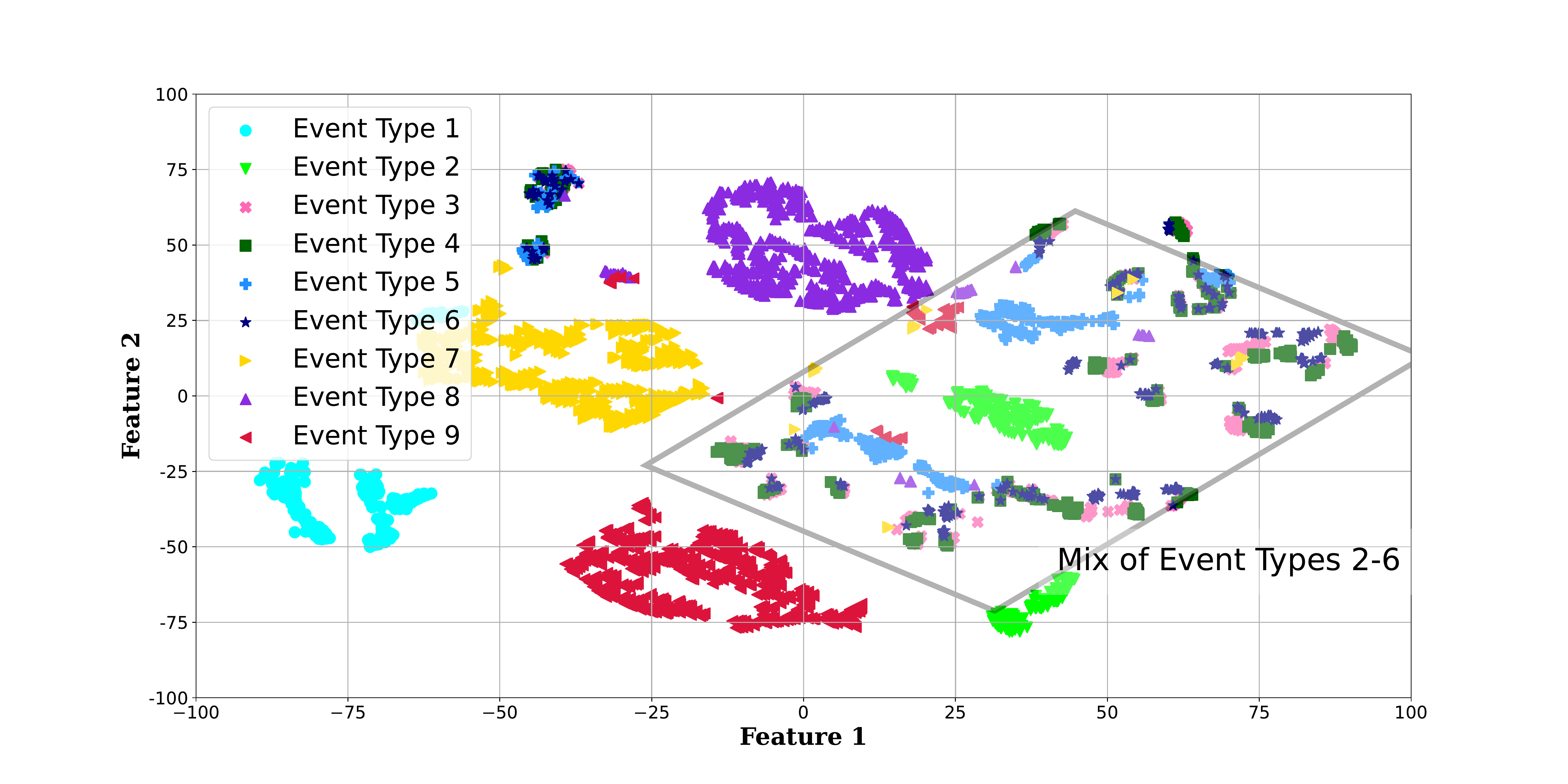}}
    
    \subfloat[GraphPMU, ARI = 0.720]{\label{sublable3}\includegraphics[trim={3.1cm 1cm 4cm 3cm},clip,scale=0.2]{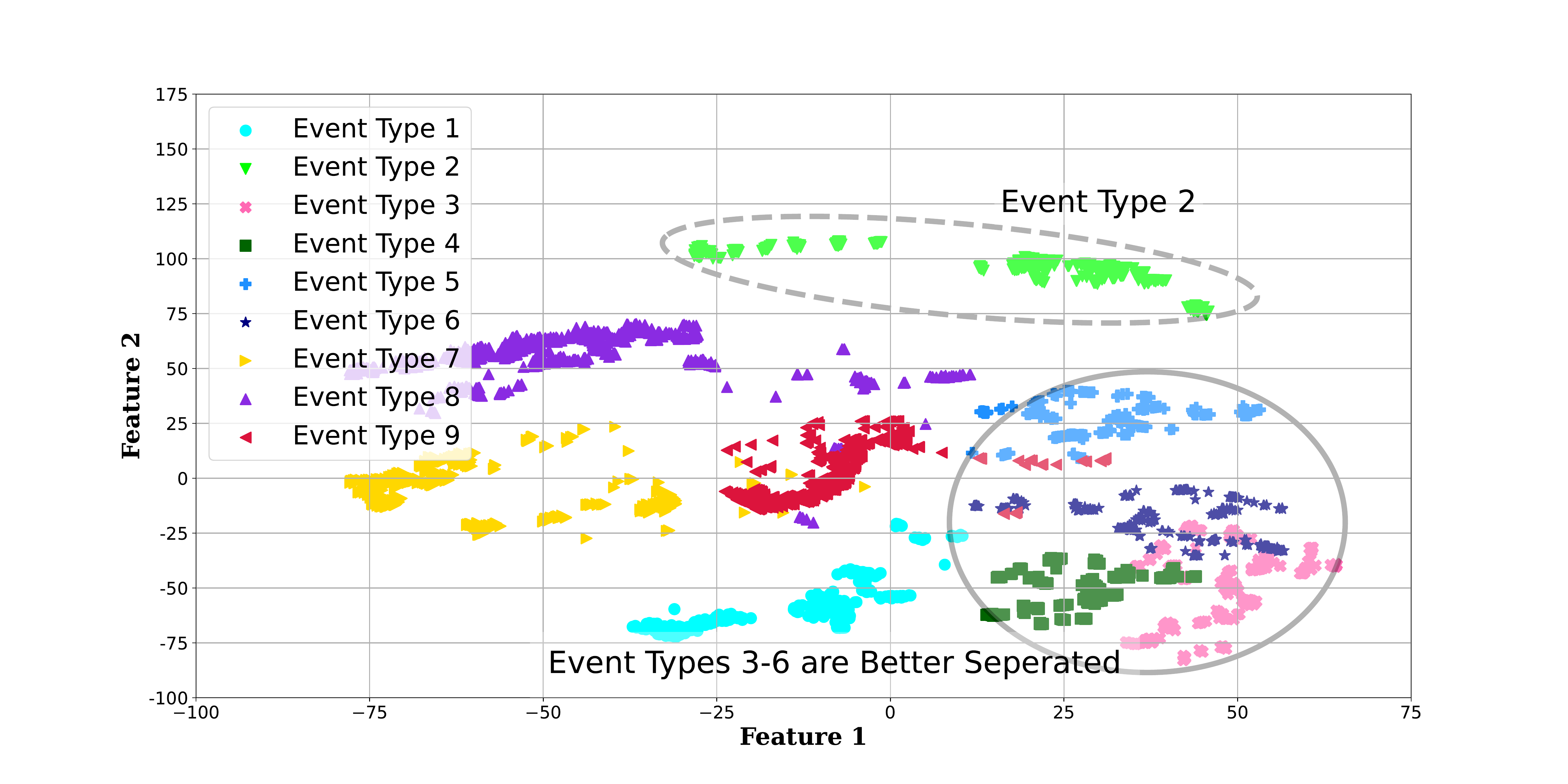}}
    \caption{The t-SNE scatter plots for the test events for three different methods based on two main features. Only four D-PMUs are available.}
    \label{fig:scatter}
    \vspace{-0.3cm}
\end{figure}

\subsection{Analysis Based on Different Types of Events} \label{subsec: clustering quality}
Fig. \ref{fig:scatter} shows the t-distributed stochastic neighbor embedding (t-SNE) scatter plot of all \emph{test} events for three methods: a) DEC; b) AED+G+NL; c) GraphPMU. 
Each point indicates one event. The shapes and colors indicate the \emph{true labels} of the nine different event types. 
%
One of the major weaknesses of the methods that do \emph{not} use graph models is their inability to properly cluster the ``smaller'' events, such as events types 3, 4, 5, and 6. For example, see the area in Fig. \ref{fig:scatter}(a) that is marked with an oval with a solid line. These four different event types are all mixed up in this area. Accordingly, the DEC method is not able to distinguish event types 3-6.

Next, consider the results in Fig. \ref{fig:scatter}(b), which are for AED+G+NL. The area that is marked with a diamond shows that AED+G+NL too is incapable of separating the ``small'' events. However, its ARI  is slightly higher than that of DEC due to the more distinctive clusters for the ``major'' event types 1, 7, 8 and 9. However, the DEC method has incorrectly split event type 2 into two separate groups of points, as we see in the two separate circles with dashed lines in Fig. \ref{fig:scatter}(b). 

GraphPMU addresses all of these shortcomings, as we can see in Fig. \ref{fig:scatter}(c). On one hand, GraphPMU tends to separate the ``major'' event types as far as possible. For example, in the dashed oval area in Fig. \ref{fig:scatter}(c), all the points for event type 2 are close to each other and away from the rest of the events. This highly improves the accuracy in clustering event type 2. 

On the other hand, GraphPMU also maintains the ``smaller'' event types reasonably away from each other. For example, in the circle area that is marked with a solid line in Fig. \ref{fig:scatter}(c), the points for event types 3, 4, 5, and 6 are separated from each other much better compared to the other figures. 
\subsection{Impact of Adding Harmonic Phasor Measurements} \label{subsec: hatmonics}
Table \ref{tab:harmonics_ARI} shows the event clustering results for AED, DEC and GraphPMU when we use not only the fundamental phasor measurements but also the harmonic phasor measurements. 
By comparing Table \ref{tab:harmonics_ARI} with Table \ref{tab:models_ARI}, we can see that the performance in event clustering has highly improved in all three methods. This is due to the more distinctive transient signatures for different event types, as we saw in Section \ref{sec:Harmonic}. 

Among the nine event types, \emph{unbalanced events} i.e., event types 3, 7 and 8, have the highest accuracy improvements. Based on Tables I and II, GraphPMU  significantly outperforms the rest of the methods, whether we only use the fundamental phasor measurements as in Table I, or we use both the fundamental and harmonic phasor measurements as in Table II. An ARI of 0.814 is very high, given that we have sensors in 4 of the 34 buses, i.e., only in 12\% of the buses. 
%
%

\begin{table}[t]

\centering
    \caption{ARI Score for GraphPMU and the Top Two Methods without Graph Models when Adding Harmonic Phasor Measurements}

    \scalebox{1.1}{
        \begin{tabular}{|c|c|}
            \hline Method & ARI \\\hline
             AED (Fundamental + Harmonics) & 0.666 \\\hline
             DEC (Fundamental + Harmonics) & 0.694 \\\hline
             \textbf{GraphPMU (Fundamental + Harmonics)} & \textbf{0.814} \\\hline
        \end{tabular}}
    \label{tab:harmonics_ARI}

\end{table}

\subsection{Impact of the Number of D-PMUs} \label{subsec: number of pmus}

\begin{figure}[b]
\begin{center}
\vspace{-0.2cm}
\includegraphics[trim={3cm 1cm 4cm 1.5cm},clip,scale=0.20]{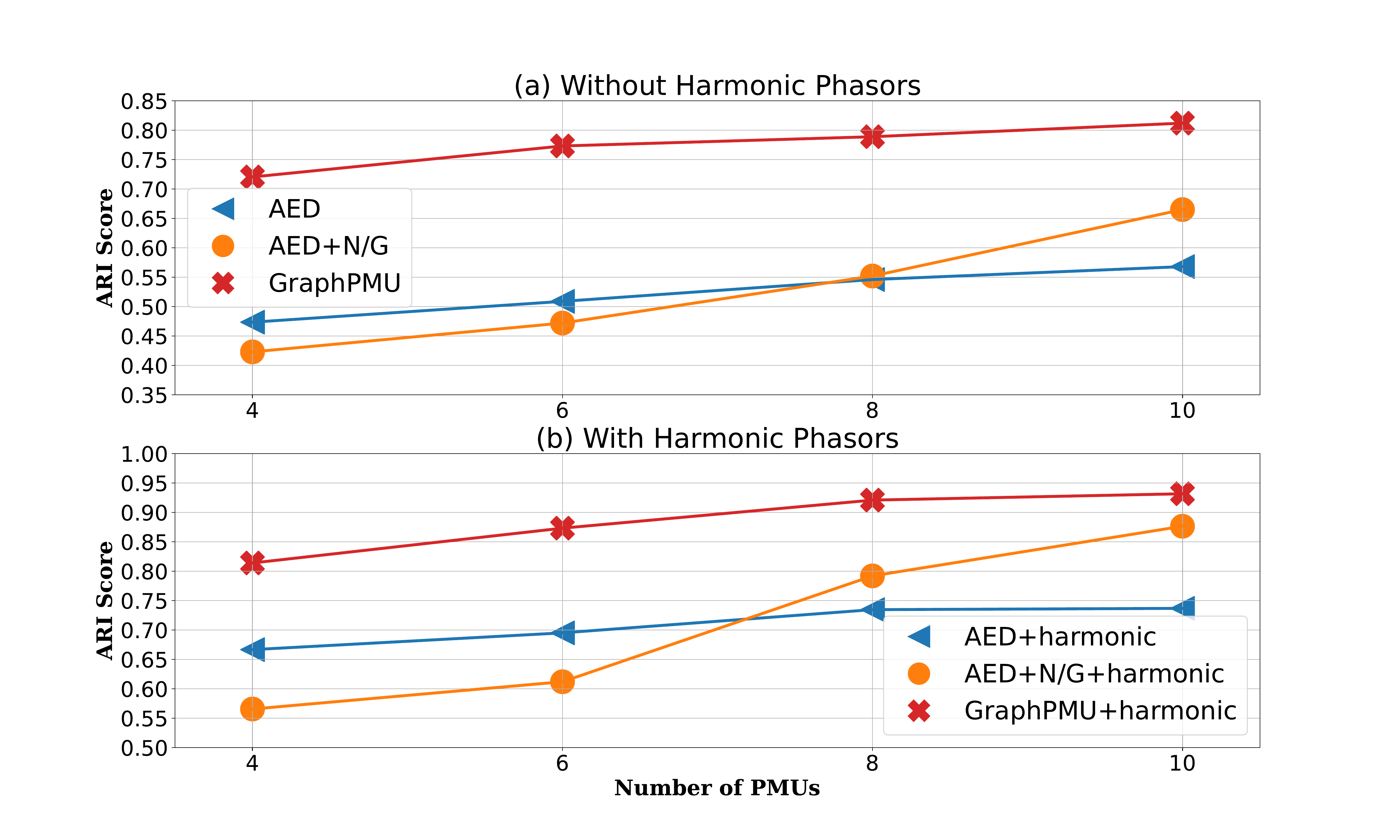}
\end{center}
\vspace{-0.3cm}
\caption{ARI scores for AED, AED+N/G and GraphPMU methods vs. the number of D-PMUs, \emph{with} and \emph{without} using harmonic synchrophasors.} \label{fig:allmodelsARI}
\vspace{-0.2cm}
\end{figure}

Fig. \ref{fig:allmodelsARI} shows the ARI scores for GraphPMU in comparison with two other  methods versus different number of available sensors. We can identify \emph{three patterns} in these figures. First, GraphPMU always outperforms the rest of the methods. Its relative superior performance is the highest when we have fewer sensors, i.e., under the \emph{locational-scarcity} conditions. Second, as we increase the number of available sensors, the overall clustering accuracy improves for \emph{all} these methods. 
Third, AED+N/G always has a worse performance than AED under severe locational-scarcity, but it surpasses AED as we increase the number of sensors. 
This is due to the fact that, AED+N/G is capable of taking advantages of the information about the network topology only when we have several sensors available. This shortcoming is addressed by GraphPMU.


\begin{figure}[t]
\begin{center}
\includegraphics[trim={3cm 0.5cm 0 3cm},clip,scale=0.203]{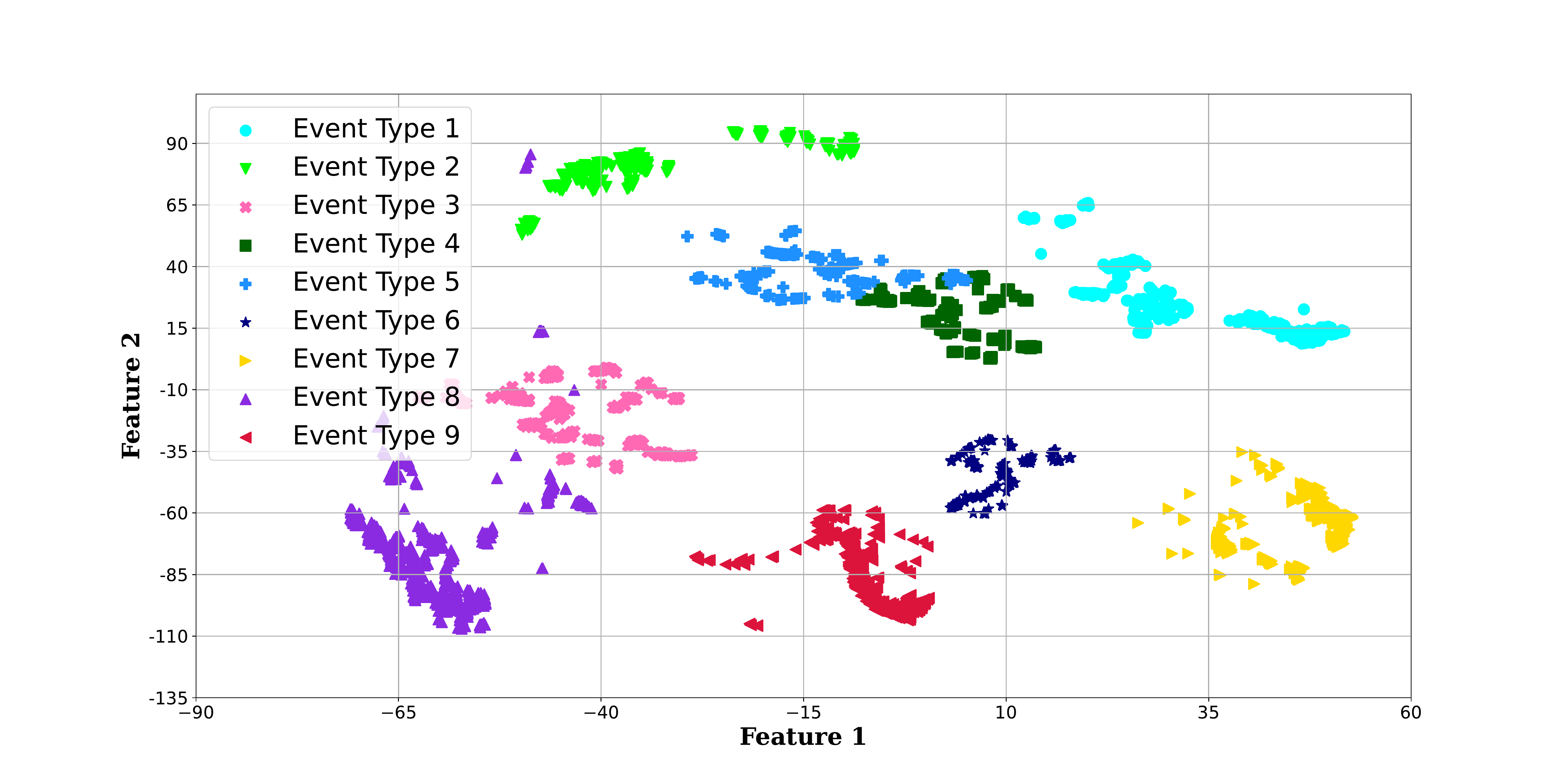}
\end{center}
\vspace{-0.3cm}
\caption{The t-SNE for GraphPMU for all test events when using 10 D-PMUs.} 
\label{fig:tsnegraph10}
\vspace{-0.05cm}
\end{figure}


In Fig. \ref{fig:allmodelsARI}(b), GraphPMU achieves a very high ARI score of 0.92 with only 8 H-PMUs in a network with 34 buses. 

Fig. \ref{fig:tsnegraph10} shows the performance of GraphPMU in clustering different types of events, when there are 10 sensors available. If we compare Fig. \ref{fig:tsnegraph10} with Fig. \ref{fig:scatter}(c), we  see that having more D-PMUs helps GraphPMU to put almost all events in correct separated clusters, for both ``major'' or ``small'' event types. 



\section{Conclusions and Future Work}
\label{conslusion}

A novel unsupervised graph-representation learning method, called GraphPMU, was proposed to cluster different types of events in power distribution systems. The proposed method does not require any prior knowledge about the events. It is solely based on the event signatures in D-PMU (and H-PMU) measurements, as well as the information about the network topology. Importantly, GraphPMU is meant to address a challenging scenario, where the phasor measurements are \emph{locationally scarce}. 
%
By conducting a comprehensive data-driven analysis, it was shown that the proper combination of \emph{topology-based}  and \emph{temporal-based} representation learnings of phasor measurements can result in very high clustering accuracy.
The results of different case studies confirmed that the proposed method outperforms the existing methods in the literature.
%
%
By using the measurements from not only fundamental but also the harmonic phasors, we further improved the clustering accuracy, particularly for unbalanced event types. 


Future work may include: 1) extending the analysis to also achieve unsupervised event location identification; 2) applying the proposed method to other sensor measurements such as synchronized waveform measurements; and 3) incorporating some additional physical information to the graph-based analysis, such as the impedance of the distribution lines. 



\bibliographystyle{IEEEtran}
\bibliography{main}

\end{document}